\documentclass[lettersize,journal,twoside]{IEEEtran}
\usepackage{amsmath,amsfonts}
\usepackage{algorithm}
\usepackage{array}
\usepackage[caption=false,font=normalsize,labelfont=sf,textfont=sf]{subfig}
\usepackage{textcomp}
\usepackage{stfloats}
\usepackage{verbatim}
\usepackage{graphicx}
\usepackage{cite}
\usepackage{caption}

\usepackage{longtable}
\usepackage{booktabs} 
\usepackage{multirow}
\usepackage{siunitx}
\usepackage{makecell}

\usepackage{algpseudocode}

\usepackage{tabularx}
\usepackage{svg}
\usepackage{subcaption}
\usepackage{float}
\usepackage{changepage}
\usepackage{soul}
\sethlcolor{yellow}
\usepackage{hyperref}
\usepackage{url}

\usepackage{colortbl}
\usepackage{xcolor}

\soulregister\cite7
\soulregister\ref7

\begin{document}

\title{Depth-PC: Sim-to-Real Transfer for Zero-Shot\\ Visual Servoing via Cross-Modal Fusion}

\author{Haoyu Zhang,
        Yang Liu,~\IEEEmembership{Student Member,~IEEE,}
        Yimu Jiang,~\IEEEmembership{Student Member,~IEEE,}
        Weiyang Lin,~\IEEEmembership{Member,~IEEE,}
        Chao Ye,~\IEEEmembership{Member,~IEEE}
        
\thanks{Manuscript received: June 26 2025; Accepted: September 1, 2025. This paper was recommended for publication by Editor Pascal Vasseur upon evaluation of the Associate Editor and Reviewers'comments. This work was supported by the National Natural Science Foundation of China under grant 62503134. \textit{(Corresponding author: Chao Ye.)}}

\thanks{The authors are with the Research Institute of Intelligent Control and Systems, Harbin Institute of Technology, Harbin 150001, China (e-mail: zhyennui@gmail.com, 24s104227@stu.hit.edu.cn,  yimujiang@stu.hit.edu.cn, wylin@hit.edu.cn, yechao@hit.edu.cn).}
\thanks{Digital Object Identifier (DOI): see top of this page.}}

\markboth{IEEE ROBOTICS AND AUTOMATION LETTERS. PREPRINT VERSION. ACCEPTED SEPTEMBER~2025}%
{\MakeLowercase{\textit{zhang et al.}}: Depth-PC: Sim-to-Real Transfer for Zero-Shot Visual Servoing via Cross-Modal Fusion}

\IEEEpubid{0000--0000/00\$00.00~\copyright~2021 IEEE}

\maketitle

\begin{abstract}
Visual servoing techniques guide robotic motion using visual information to accomplish manipulation tasks, requiring high precision and robustness against noise. Traditional methods often require prior knowledge and are susceptible to external disturbances. Learning-driven alternatives, while promising, frequently struggle with the scarcity of training data and fall short in generalization. To address these challenges, we propose \textit{Depth-PC}, a novel visual servoing framework that leverages decoupled simulation-based training from real-world inference, achieving zero-shot Sim2Real transfer for servo tasks. To exploit spatial and geometric information of \textit{depth} and \textit{point cloud} features, we introduce cross-modal feature fusion, a first in servo tasks, followed by a dedicated Graph Neural Network to establish keypoint correspondences. Through simulation and real-world experiments, our approach demonstrates superior convergence basin and accuracy compared to SOTA methods, fulfilling the requirements for robotic servo tasks while enabling zero-shot Sim2Real transfer. In addition to the enhancements achieved with our proposed framework, we have also demonstrated the effectiveness of cross-modality feature fusion within the realm of servo tasks. Code is available at https://github.com/3nnui/Depth-PC. 
\end{abstract}


\section{Introduction}
\IEEEPARstart{G}{iven} Given the current image, Visual Servoing (VS) \cite{1,2} aims to guide the movement of robots for precise relative position control. To achieve this, robots need to automatically adjust their actions for desired control outcomes. Owing to its versatility and precision, VS has demonstrated significant potential across a wide range of applications, including industrial automation \cite{3}, visual navigation \cite{4}, and autonomous 
aerial vehicles \cite{5}. Previous methods can be roughly categorized into two branches according to learning space within closed-loop control: transitioning from image-based visual servoing (IBVS) to position-based visual servoing (PBVS).

Concretely, IBVS-based solutions \cite{1,8} control robots by minimizing feature errors explicitly in the image space. This methodology can be traced back to keypoint correspondence between current and desired images, which promises robustness against the relative motion. On the other hand, PBVS controllers \cite{2,10} incorporate camera calibration information for target's location, which avoid explicit error correspondence and facilitate control within 3D space. 

\begin{figure}[t]
    \centering
    \includegraphics[scale=0.54]{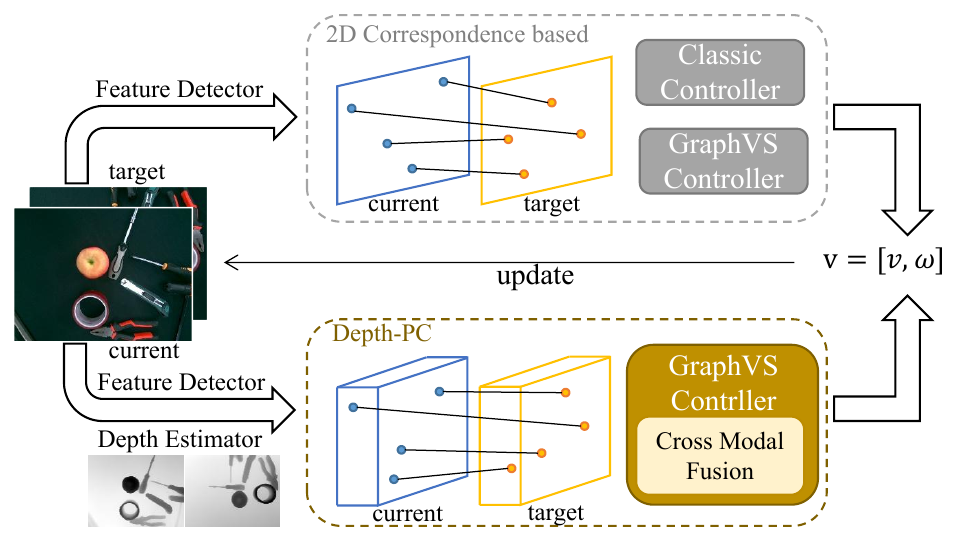}
    \caption{Comparative visualization between visual servoing methods based on 2D correspondence and ours. In the inference stage, apart from the feature detector, We utilize the off-shelf depth estimator to inject spatial and geometric cues into the system.}
    \label{fig:1_Teaser_img}
    \vspace{-15pt}
\end{figure}

IBVS controllers are limited by a small convergence region and susceptibility to local minima, while PBVS ones are prone to interference from relative motion \cite{13}. Recent works such as CNS \cite{21} utilize explicit correspondence with the neural policy to combine the both advantages. Additionally, NUVS \cite{NUVS} is designed to estimate the calibration embedding via point correspondence. While highly effective, they are highly dependent on reliable matching of hand-crafted 2D features, as shown in Fig. \ref{fig:1_Teaser_img}. Crucially, during the transfer from simulation to real-world, such 2D features face cross-domain generalization challenges. The matching accuracy and stability can degrade significantly due to environmental discrepancies. Pure 2D correspondence features prove inadequate in resolving 3D ambiguities [35], a limitation that even depth-aware methods [34], [35], [36] struggle to overcome, as their reliance on noisy depth maps often compromises geometric fidelity. This situation leads to a critical question: \textit{Given the inherent inadequacies of 2D features in bridging the Sim2Real gap and enabling precise visual servoing in 3D world, can and how can leveraging 3D features offer a more robust solution?}

\IEEEpubidadjcol 
In this research, we propose Depth-PC, a zero-shot Sim2Real framework for visual servoing. Our objective is to perform servo tasks without the need for additional training to adapt to a variety of scenarios in a zero-shot manner. We achieve this by training our network on randomized simulation data, while adopting a distinct approach for inference. Besides, instead of raw depth maps, our framework leverages an off-the-shelf depth estimator to construct robust 3D representations that generalize across Sim2Real domains.

\begin{figure*}[htpb]
    \centering 
    \includegraphics[scale=0.51]{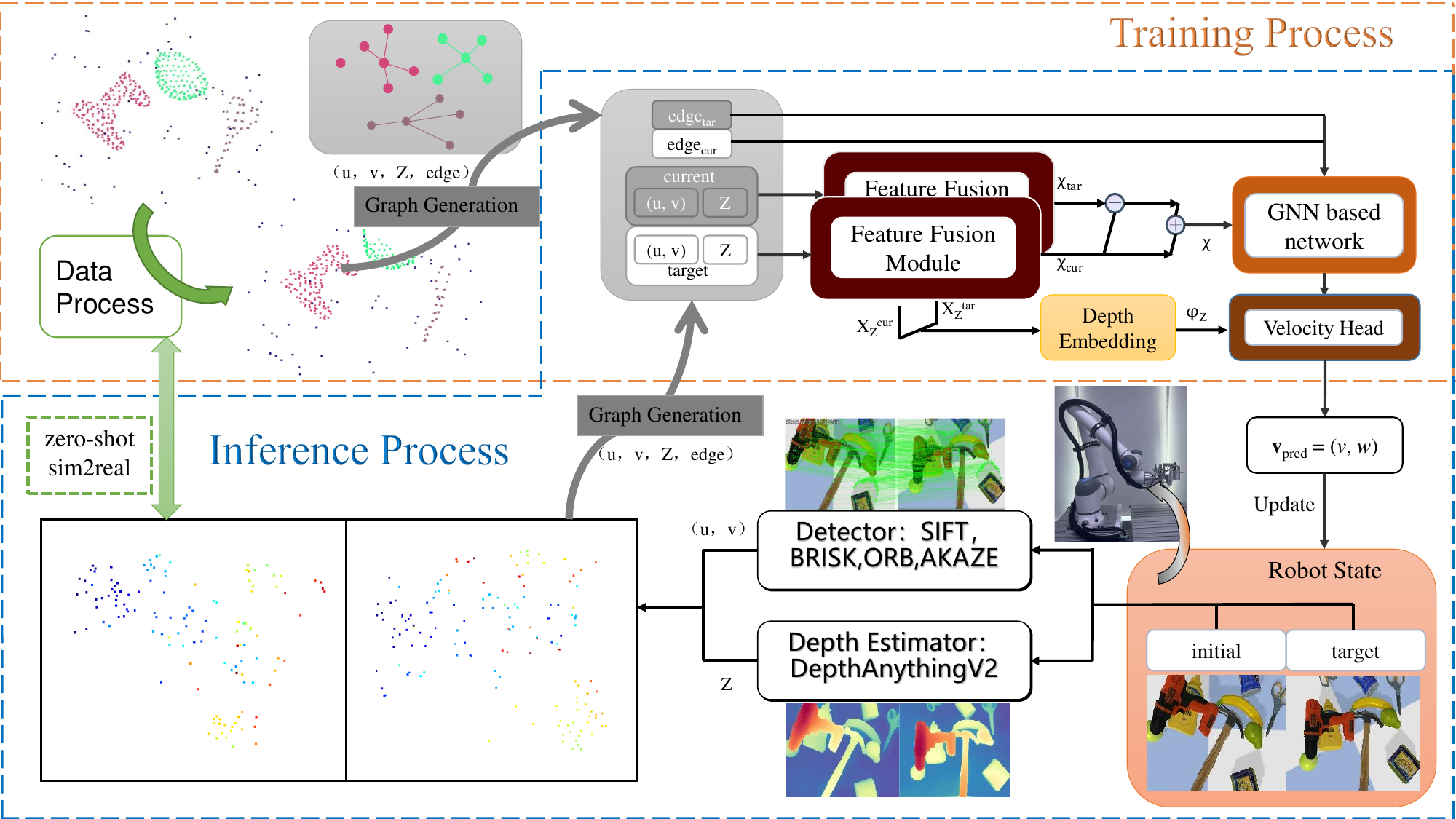}
    \caption{Overview of our framework. The proposed framework is separated into training and inference components. During training, processed object point clouds are fed into a Graph Neural Network (GNN)\cite{28}. In the inference phase, the inputs are derived from point clouds and simulation-aligned depth features generated by a relative depth estimator. The fused features are then used to construct graph relationships, which produce 6-DOF velocity to guide the robot in executing servo tasks.}
    \label{fig:2_Framework}
    \vspace{-15pt}
\end{figure*}

In summary, the contributions of our work are as follows: 

\begin{adjustwidth}{-0.3cm}{0cm}
    \begin{itemize}
        \item We have developed a novel framework for servo tasks, which achieves Sim2Real effectiveness by separating trainning and inference processes, and aligns the simulation space with the real world, thereby extending generalizability to other scenarios in a zero-shot way.
        \item To our knowledge, we are the first to introduce cross-modal feature fusion into servo tasks by aligning feature information across diverse domains  to bridge the gap between simulation and real-world scenarios.
        \item Our method's performance has been validated through zero-shot assessments conducted in both simulation environments and real-world scenes, demonstrating superior Sim2Real transfer capability.
    \end{itemize}
\end{adjustwidth}

\vspace{-7pt}
\section{RELATED WORK}

\subsection{Traditional Visual Servoing}

Traditional visual servoing methodologies \cite{1,2} include IBVS, PBVS, and Hybrid Visual Servoing. IBVS methods leveraging geometric features directly derived from images \cite{29} face challenges such as unpredictable motion paths, potential Jacobian matrix singularities, local minima, and complications in feature extraction. On the other hand, PBVS \cite{30} relies on 3D position data as visual cues, ensuring global asymptotic stability. But it necessitates prior knowledge like accurate camera parameters or 3D models of the target objects for pose estimation. Hybrid VS integrates image-level insights with 3D reconstruction from dual viewpoints of a rigid object, enabling the decoupling of rotation and translation parameters\cite{31}. Nevertheless, this approach increases system complexity, posing challenges to the effective integration and optimization of these features.

\vspace{-5pt}
\subsection{Learning-based Visual Servoing}

Deep learning models excel in enhancing generalizability, precision, and adaptability. One pathway involves predicting the relative 6D orientation of objects, while the alternative strategy predicts the velocity of the end-effector to achieve the desired state. Prior methods for 6D pose estimation often rely on instance-specific or category-level frameworks \cite{15,34}, necessitating CAD models of particular objects, and are limited to those specific cases. Generalization to novel instances or categories demands additional techniques, like NeRF \cite{33}. To counter these limitations, alternative approaches leverage depth information or point clouds, enriching the pose estimation with semantic cues. After initial pose estimation from reference perspectives, Gen6D \cite{35} leverages a feature volume and networks for the refinement of estimated poses. OnePose \cite{36} employ SfM \cite{38} to generate a sparse point cloud and ascertain the poses. Yet, these methods require substantial annotated data, leading to high costs. Balancing real-time performance and precision in robotic manipulation intensifies these challenges.

\vspace{-3pt}
Meanwhile, methods focusing on image features for velocity feedback have emerged. KOVIS \cite{19} exemplifies this by training within simulation scenarios, harnessing supervised segmentation and depth information to enable Sim2Real transfer. Other techniques \cite{14,17,39} focus on mapping current and target features to a latent space and utilizing a MLP to output commands. While robust against intricate environments, these methods may not generalize seamlessly to new settings or adeptly manage rotation variances. CNS \cite{21} enhances generalization by employing randomized simulated point clouds to forge feature correspondence, yet its performance wavers in noisy environments due to the absence of spatial and geometric insights. Inspired by this control baseline, our research further elevates the robot's perceptual capabilities into the three-dimensional realm.

\vspace{-5pt}
\subsection{Handling Depth in Visual Servoing}

Some works have delved into leveraging depth maps to construct servo systems. One such approach \cite{2016depth} utilizes depth maps of the robot to directly estimate the joint positions. Another method \cite{2019depth} uses visual features from color edges and learned depth keypoints to predict robotic states from depth images. In addition to optical flow features, the approach \cite{2020dfvs} introduces depth information and employs it for multi-perspective depth rectification. Nevertheless, these methodologies grapple with inherent issues such as inaccuracy of depth in complex settings, limited convergence basins, and a lack of robust generalization across diverse scenarios. Essentially, aiming to address the depth estimation problem, they do not offer tailored solutions for the intricacies of robotic control.

\section{Method}

To endow data with enriched spatial and geometric properties, and align features of simulation environment with those of real world, our approach employs two strategies: 1). leveraging the "Hidden Points Removal" \cite{43} operator along with normalized depth information; 2). crafting a feature fusion module designed to align and fuse depth data seamlessly with feature points. Besides, we integrate a relative depth estimation model \cite{42} to facilitate the correspondence of depth features in the inference phase. 

\vspace{-5pt}
\subsection{Data Processing}\label{sec1:Data Processing}

Our methodology generates data with spatial and geometric information, an attribute absent in CNS data \cite{21}. We integrate models \footnote{\url{https://github.com/c-yiting/pybullet-URDF-models}} into the simulation environment with randomized numbers of point clouds. Specifically, we use the objects in the YCB-Video dataset for testing, while the point clouds of the remaining objects are used for training.

\begin{algorithm}[H]
    \caption{Pipeline of Data Generation and Preprocessing}
    \label{alg:data_process}
    \begin{algorithmic}[1]
        \State Voxel downsample\footnote{\url{https://www.open3d.org/}} $N_{c}$ 3D model point clouds $ P_{i}^{f} =\{ V(P_{i}^{o} \in P^{o}) \mid 1\leq i \leq N_{c} \}$; 
        \State Given \(\it{N}\) points and \(n_{i}\) points in the \(\mathit{i}\)-th cluster, randomly select points with randomized centroids and rotations to form $\mathcal{P}_{i} = \{ p_{j} \in P_{i}^{f} \mid 1\leq j \leq n_{i} \}$; 
        \State Generate camera poses ${}_{w}^{c}T_{i}$ and ${}_{w}^{c}T_{t}$ with perturbations;
        \State Obtain the essential data points through Hidden Points Removal operator: $\hat{\mathcal{P}}_{i} = \{ HPR({{p}_{j}}) \mid 1\leq j \leq n_{i}$ \};
        \State Project points to 2D camera plane as keypoints $\hat{{P}}$ and normalize with depth $Z$;
        \State Augment data: randomly select mismatched points, omit some points, and add uniform noise;
        \State Generate graph correspondence, distinguish inter- and intra-cluster connections, as the network input.
    \end{algorithmic}
\end{algorithm}

Within this context, we define a cylindrical spatial region $\mathit{S}(r, h)$ to enclose the input data from the 3D models' point clouds $P^{o}$. The pipeline of data generation and preprocessing is shown in Algorithm \ref{alg:data_process}.

\footnotetext[2]{\url{https://www.open3d.org/}}

Similar to CNS \cite{21}, we generate initial camera pose ${}_{w}^{c} T_{i}$ and target camera pose ${}_{w}^{c} T_{t}$ within a hemisphere based on the scene's center, ensuring that their viewing directions are oriented toward the centroid. Additionally, some perturbations are added to simulate real-world robotic uncertainties. Leveraging these poses as pivotal reference postions, we then apply the Hidden Points Removal $\mathit{HPR(\cdot)}$ operator to emulate the complexities of occlusions and visibility constraints in the real world, as well as further reduce the number of data points.

\begin{figure}[htpb]  
    \centering  
    \includegraphics[scale=0.4]{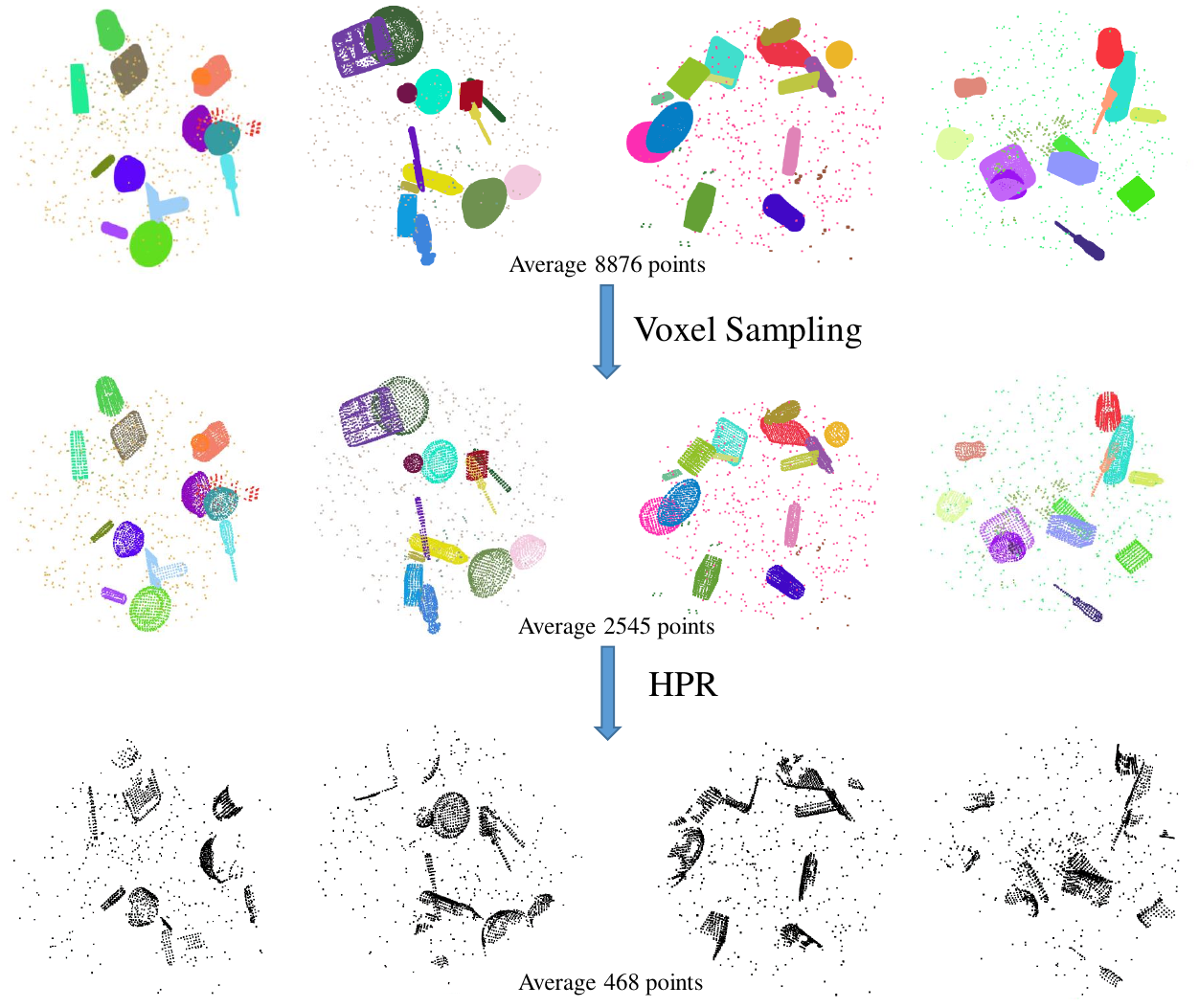}
    \captionsetup{skip=-2pt} 
    \caption{The illustration shows the process of data generation. Point clouds of each object are colored to distinguish their differences, with average number of points in the middle. It indicates that our data itself contains certain spatial and geometric information.}
    \label{fig:3_Data_Generation} 
\end{figure} 

The whole process is illustrated in Fig. \ref{fig:3_Data_Generation}. This approach strikes a balance between the volume of data and the maximal extraction of spatial and geometric information from the data, a strategic decision that aims to minimize the disparity in the distribution of point clouds between the training dataset and the real-world scenario.

After the sampled points undergo projection and augmentation, graph correspondence is established among the points (node embeddings) with two types of connections included. Inter-cluster connections link centroids of different point clusters to model inter-group spatial relationships, and intra-cluster connections form an adjacency matrix by connecting points within the same cluster. They are are obtained by the Affinity-Propagation algorithm \cite{frey2007clustering} with random parameters like CNS. The resulting graph—encoded by node and edge embeddings—serves as the network input.

\vspace{-5pt}
\subsection{Feature Fusion Module}\label{sec2:Feature Fusion Module}

In order to enhance the spatial and geometric richness of the sampled points, we augment the data with the depth information, which is the normalized distance from each point to the camera plane.
We begin by mapping the normalized 2D feature points \(\hat{\mathit{P}}_{\mathrm{norm}}\) and depth information \(Z_{\mathrm{norm}}\) into a unified feature space through FAL (Feature Alignment Layer), yielding feature points map and feature depth map: 
\begin{equation}
    \begin{aligned}
        {\mathit{X}}_{\mathrm{pos}} = \mathrm{FAL}(\hat{\mathit{P}}_{\mathrm{norm}}) , {\mathit{X}}_{\mathrm{Z}} = \mathrm{FAL}(\mathit{{Z}_{\mathrm{norm}}}) 
    \end{aligned}
\end{equation}

\begin{figure}[H]       
    \centering
    \includegraphics[scale=0.4]{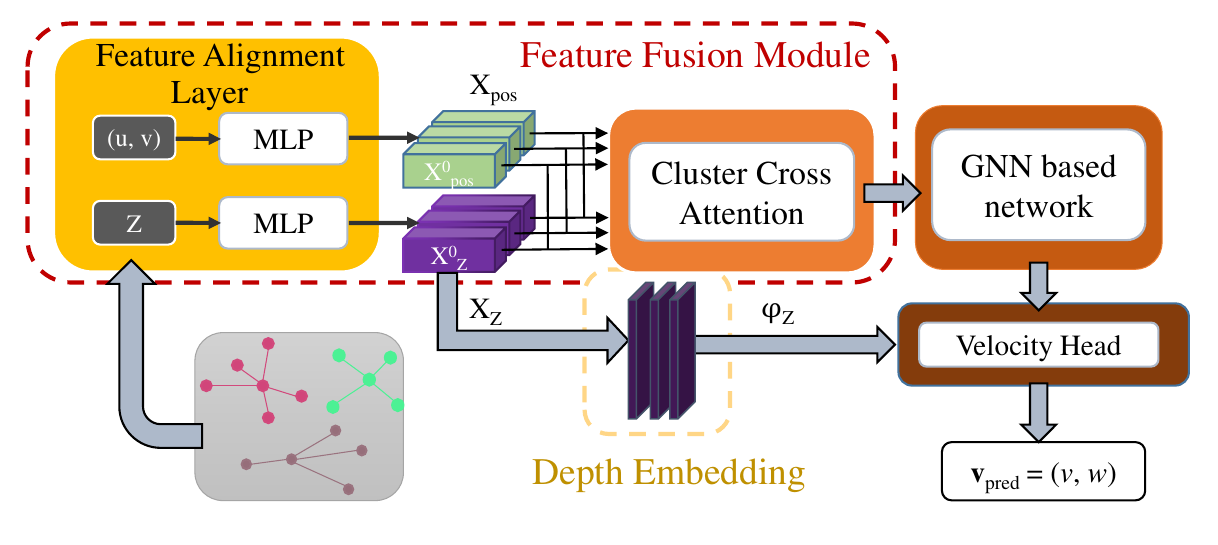}
    \caption{Our feature fusion module are mainly made of Feature Alignment Layer and Cluster Cross Attention. The former aligns two feature spaces, while the latter fuses the positional features of each cluster with the depth features of all points. Followed by GNN and Velocity Head, we finally get the 6-DOF velocity.}
    \label{fig:4_depth_emb}
\end{figure}

Despite its efficacy, Cross Attention suffers from high training costs, motivating us to introduce Cluster Cross Attention to enchance cross-modality interaction, as shown in Fig. \ref{fig:4_depth_emb}. 
Specifically, for \(\mathit{i}\)-th cluster, feature fusion involves calculating the attention score matrix:
\begin{equation}
    \begin{aligned}
        \mathit{score}^{\mathit{i}} = {\mathit{X}}_{\mathrm{pos}}^{\mathit{i}} 
        \cdot {\mathit{X}}_{\mathrm{Z}}^\top \in \mathbf{R}^{n_{i} \times N_{d}} 
    \end{aligned}
\end{equation}
where \(\mathit{n_{i}}\) is the number of points in \(\mathit{i}\)-th cluster and $\mathit{N_d}$ is feature dim. 

Subsequently, Softmax is applied to each row of the \textit{scores} matrix to derive the attention matrix that represents the focus of depth information on coordinate information; similarly, column-wise Softmax generates the coordinate-to-depth attention matrix:
\begin{equation}
    \begin{aligned}
        \mathcal{A}_{Z}^{\mathit{i}} = \mathrm{Softmax}_{r}(\mathit{score}^{\mathit{i}}) ,
        \mathcal{A}_{pos}^{\mathit{i}} = \mathrm{Softmax}_{c}(\mathit{score}^{\mathit{i}})
    \end{aligned}
\end{equation}

Ultimately, the updated fused features are obtained as:
\begin{equation}
    \begin{aligned}
        \mathcal{X}^{\mathit{i}} = [
        \mathcal{A}_{Z}^{\mathit{i}} \cdot {\mathit{X}}_{\mathrm{Z}}^{\mathit{i}},    \mathcal{A}_{pos}^{\mathit{i}} \cdot {\mathit{X}}_{\mathrm{pos}}^{\mathit{i}}]
    \end{aligned}
\end{equation}

Besides, the depth branch of the fused features is fed into MLP, considered as depth embeddings \({\varphi}_Z\).

\subsection{Training and Inference}\label{sec3:Training and Inference}

\textbf{Training Process} Our training process is presented within the orange dashed box depicted in Fig. \ref{fig:2_Framework}. Once acquiring the data from section \ref{sec1:Data Processing}, we feed the node embeddings of the established graph correspondences into the Feature Fusion Module (FFM) to perform the fusion of two features. Subsequently, the resulting node embeddings and the edge embeddings are processed by the dedicated Graph Neural Network (GNN) \cite{28} which aims to minimize errors between current and target states. The velocity head, as a decoder, yields the 6 DOF velocity of the end-effector. Furthermore, we integrate depth embeddings into MLP, leveraging its robust feature extraction capabilities to refine the velocity decoupled from distance. This approach injects spatial and geometric knowledge into the model, thereby extending its adaptability to diverse environmental settings. 

Similarly, we also use PBVS \cite{10} to supervise the velocity which is decoupled from the scene distance prior $\mathit{d}^{gt}$. The ground truth velocity can be represented as: $\mathbf{v}^{gt}_{d}=[\mathit{v}^{gt}/\mathit{d}^{gt} ; \omega^{gt}]$ and the loss function can be formulated with the predicted velocity $\mathbf{v}^{pred}_{d}$ as follows:
\begin{equation}
    \begin{aligned}
    {\mathcal{L}_\mathrm{dir} = 1 - \mathrm{CosineSimilarity}(\mathbf{v}^{pred}_{d}, \mathbf{v}^{gt}_{d}}) \\
    {\mathcal{L}_\mathrm{norm} = \mathrm{MSE}(\mathbf{v}^{pred}_{d}, \mathcal{T}^{-1}(\|\mathbf{v}^{gt}_{d}\|_2))}
    \end{aligned}
\end{equation}
where $\mathcal{T}(\cdot) = 1 + \mathrm{ELU}(\cdot)$. The total loss can  be formulated as $\mathcal{L}_\mathrm{total}=\mathcal{L}_\mathrm{dir}+0.1\mathcal{L}_\mathrm{norm}$ and the predicted velocity is multiplied by a distance coefficient to control the robot's motion.

\begin{figure}[thbp]
    \centering
    \begin{tabular}{@{\extracolsep{\fill}}c@{}c@{\extracolsep{\fill}}}
            \includegraphics[width=0.48\linewidth]{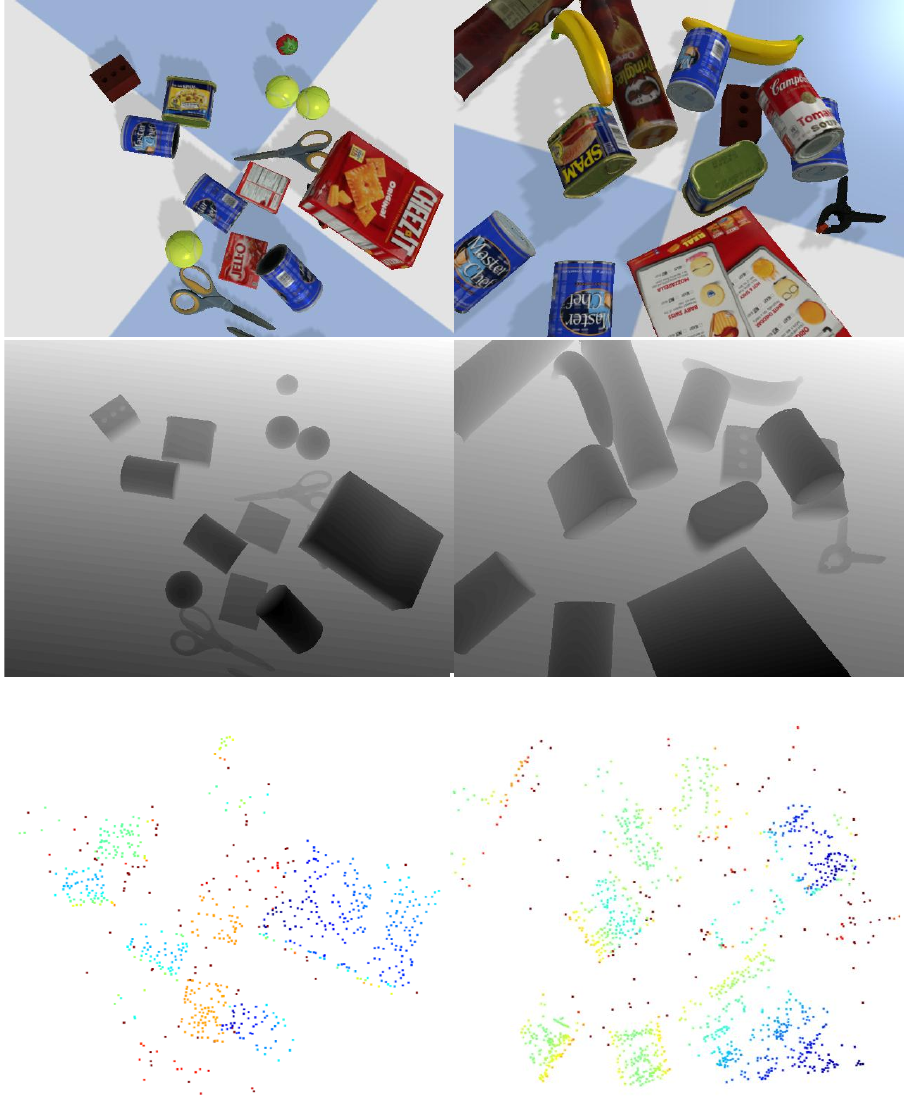} &
            \includegraphics[width=0.48\linewidth]{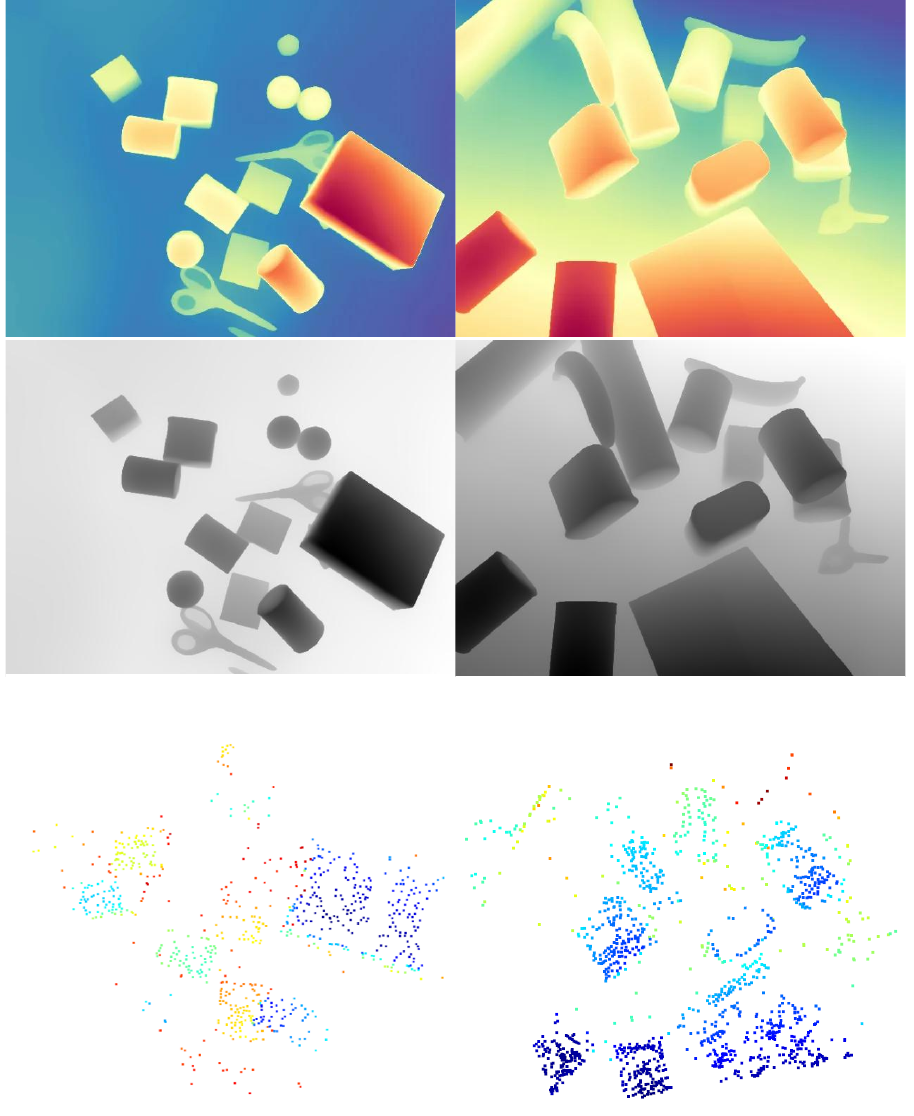}\\
            (a).simulation results & (b).DepthAnythingV2 results\\
    \end{tabular}
    \caption{Visualization of depth results. The figure on the left illustrates the depth maps rendered within the simulation environment, along with keypoint clouds detected; the figure on the right presents frames and keypoint clouds processed by the depth estimator. It is obvious that depth distributions of point clouds between them share a high degree of similarity.}
    \label{fig:5_depth_sim}
\end{figure}

\textbf{Inference Process} Due to the limitations of 3D cameras in reality, depth information is only effective within a certain range, and also greatly affected by external noise, as shown in Fig. \ref{fig:9_different_depths}. Meanwhile, DINOv2 \cite{27} demonstrates an exceptional ability to mine semantic information from images, leading to remarkable outcomes in relative depth estimation. This capability complements our normalized depth data, enabling the seamless alignment of semantic information between the simulation environment and real world, as shown in Fig. \ref{fig:5_depth_sim}. Consequently, we have integrated pre-trained DepthAnythingV2 \cite{42} (DINOv2-based) in our experiments to extract relative depth information, a critical component of our input data. And the normalized relative depth information corresponds to $Z_{\mathrm{norm}}$ mentioned in section \ref{sec2:Feature Fusion Module}. Additionally, feature points, corresponding to $\hat{\mathit{P}}_{\mathrm{norm}}$, can be obtained from any arbitrary keypoint detector, offering flexibility in our approach.

\begin{table*}[htbp]
    \centering
    \caption{We conduct experiments in \(E_\mathrm{YCB}\) in a zero-shot way and consider the servo task to be successful when the mean points error is within the threshold and the state is maintained for 20 steps. The bolded ones indicate the best performance, and the underlined ones come next.}
    \label{tab:1_Eycb}
    \begin{tabular}{cccccccc}
        \toprule
            Detector & Controller & SR(\%) & TE(mm) & RE(\(^{\circ}\)) & TS(0.04s) \\  
        \midrule
            RAFT & IBVS \cite{1} & \textbf{100}  & $2.336 \pm 4.642$ & $0.164 \pm 0.302$ & $358.3 \pm 94.3$ \\ 
        \midrule
             ViT & IBVS \cite{1} & $78.67 \pm 39.43$ & $1.762 \pm 1.273$ & $0.324 \pm 0.292$ & $375.9 \pm 93.6$ \\ 
        \midrule
        \multirow{3}{*}{SIFT\cite{lowe1999object}} 
            & IBVS & $94.00 \pm 23.75$ & $6.633 \pm 7.472$ & $0.550 \pm 0.618$ & $231.2 \pm 92.4$ \\ 
            & CNS \cite{21}  & $96.00 \pm 19.60$ & $2.486 \pm 3.079$ & $0.186 \pm 0.242$ & $357.4 \pm 145.4$ \\ 
            & Depth-PC(ours) & $98.67 \pm 11.47$ & $\underline{1.466} \pm 2.452$ & $\textbf{0.117} \pm 0.187$ & $216.4 \pm 78.8$ \\ 
        \midrule
        \multirow{3}{*}{ORB\cite{rublee2011orb}} 
            & IBVS \cite{1} & $74.00 \pm 43.86$ & $4.975 \pm 5.919$ & $0.425 \pm 0.540$ & $224.7 \pm 45.0$ \\ 
            & CNS \cite{21}  & $88.00 \pm 32.50$ & $3.839 \pm 3.567$ & $0.348 \pm 0.227$ & $365.8 \pm 169.2$ \\ 
            & Depth-PC(ours) & $96.67 \pm 17.95$ & $3.100 \pm 4.438$ & $0.250 \pm 0.321$ & $242.6 \pm 129.9$ \\ 
        \midrule
        \multirow{3}{*}{BRISK\cite{leutenegger2011brisk}} 
            & IBVS \cite{1} & $90.67 \pm 29.09$ & $3.768 \pm 4.64$ & $0.307 \pm 0.367$ & $235.9 \pm 59.1$ \\ 
            & CNS \cite{21}  & $98.67 \pm 11.47$ & $3.619 \pm 4.901$ & $0.298 \pm 0.321$ & $539.0 \pm 124.3$ \\ 
            & Depth-PC(ours) & $99.33 \pm 8.14$ & $1.813 \pm 1.612$ & $0.140 \pm 0.111$ & $245.5 \pm 135.2$ \\ 
        \midrule
        \multirow{3}{*}{SuperGlue\cite{sarlin2020superglue}} 
            & IBVS \cite{1} & $70.00 \pm 45.83$ & $3.121 \pm 5.431$ & $0.242 \pm 0.409$ & $\underline{187.8} \pm 24.6$ \\ 
            & CNS \cite{21}  & $65.12 \pm 47.66$ & $4.305 \pm 5.573$ & $0.355 \pm 0.377$ & $276.1 \pm 154.6$ \\ 
            & Depth-PC(ours) & $76.67 \pm 42.30$ & $3.622 \pm 4.007$ & $0.273 \pm 0.264$ & $196.5 \pm 108.3$ \\ 
        \midrule
        \multirow{3}{*}{AKAZE\cite{AKAZE}} 
            & IBVS \cite{1} & $85.33 \pm 35.38$ & $1.602 \pm 1.258$ & $0.136 \pm 0.109$ & $253.9 \pm 42.2$ \\ 
            & CNS \cite{21} & $\textbf{100}$ & $1.592 \pm 1.328$ & $0.134 \pm 0.103$ & $207.8 \pm 61.1$ \\ 
            & Depth-PC(ours) & \textbf{100} & $\textbf{1.421} \pm 1.284$ & $\underline{0.125} \pm 0.093$ & $\textbf{178.6} \pm 40.0$ \\ 
        \bottomrule
    \end{tabular}
    \vspace{-15pt}
\end{table*}

\section{EXPERIMENTS}

Our experimental environments are divided into two parts: the simulation environment and the real-world environment. In both cases, we deploy our well-trained model to perform servo tasks in scenes where we randomly sample the initial and target poses using the same method described in \ref{sec1:Data Processing}. And images from the camera serve as input data for the network to directly predict the velocity of the robot's end-effector in a zero-shot way. The evaluation metrics include success rate (SR), translation error (TE), rotation error (RE), time steps (TS) and mean total time cost on each run (mTT). 

Baseline: (1). IBVS \cite{1} directly controls robot motion by minimizing the pixel errors of keypoints through the average of current and desired Jacobian matrix with depth corresponding to $\mathit{d}^{gt}$; (2). IBVS-related methods merely replace the extracted feature with classic features (SIFT \cite{lowe1999object}, ORB \cite{rublee2011orb} e.g.), optical flow features (RAFT \cite{raft}), or features from Vision Transformer (ViT-VS \cite{scherl2025vit}); (3). CNS \cite{21} is only trained on randomized points in 3D space and encodes the keypoints correspondence into a graph followed the same GNN and decoder as ours.

\vspace{-5pt}
\subsection{Simulation Environment}

In the first scenario \(E_\mathrm{YCB}\) rendered by PyBullet engine, we perform a comparative analysis with CNS \cite{21} and IBVS \cite{1} by introducing 3D models from YCB-Video dataset \cite{15}. The scenes can be categorized into three types—S, M, L—based on the positional and rotational deviation. Each category undergoes 50 iterative tests. We consider the servo task successful if the final rotation error is less than \(3^{\circ} \) and the translational error is within 3 cm.

The statistics in TABLE \ref{tab:1_Eycb} clearly demonstrate a significant 15\% improvement in convergence speed and superior accuracy. Even with identical feature extractors (e.g., SIFT, ORB), our approach still delivers markedly better performance—excelling not only in efficiency but also in greater robustness across diverse feature characteristics, thereby ensuring stable and reliable results.

\begin{figure}[htpb]       
    \centering
    \includegraphics[scale=0.5]{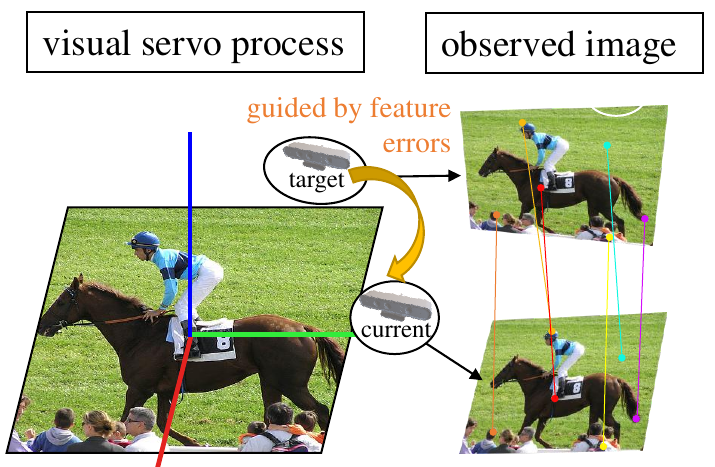}
    \caption{Illustration of servo processes in $E_\mathrm{image}$, where the images are subject to affine transformation to simulate different poses of the robots.} 
    \label{fig:6_image_voc}
    \vspace{-10pt}
\end{figure}

For the second part, according to studies denoted by \cite{21}, \cite{14} and \cite{39}, we adopt a similar approach of utilizing 2D images to construct task environment \(E_\mathrm{image}\) as shown in Fig. \ref{fig:6_image_voc}. Data augmentation is applied to these images during training for \cite{14} and \cite{39}, including the introduction of occlusions, to simulate a more complex visual scene. In the inference phase, models analyze perspectives from varied camera poses to predict actions accurately.

Due to the specific training on certain datasets, methods such as \cite{14} \cite{39} have shown some effectiveness for specific datasets. Apart from the backbone, these methods relied solely on a simple MLP to predict the output velocities. However, by capitalizing on the broad generalization abilities of \cite{21} and incorporating only the relative depth cues from images, we have managed to sidestep the constraints of varying scenes, thus attaining performance that is on par with existing SOTA as indicated in TABLE \ref{tab:2_Eimage}.

\begin{table}[htbp]
    \centering
    \caption{Experiments in \(E_\mathrm{image}\). The test pictures come from VOC2012$^1$, observed from different initial and target poses, with AKAZE as the feature detector.}
    \label{tab:2_Eimage}
    \begin{tabular}{ccccc}
        \toprule
            Method&SR(\%)&TE(mm)&RE(\(^{\circ}\))&mTT(ms)\\  
        \midrule
            AlexNet-based\cite{14}&93.20&28.854&2.365&401.0\\
            Siame-se(3)\cite{39}&92.67&54.084&6.758&401.0\\
            CNS\cite{21}&\textbf{100}&0.708&\textbf{0.037}&32.32\\
            Depth-PC(ours)&99.33&\textbf{0.626}&0.124&\textbf{32.06} \\
        \bottomrule
    \end{tabular}
\end{table}

\footnotetext[1]{\url{http://host.robots.ox.ac.uk/pascal/VOC/voc2012/}}

\begin{figure*}[htpb]
    \centering  
    \includegraphics[width=\linewidth]{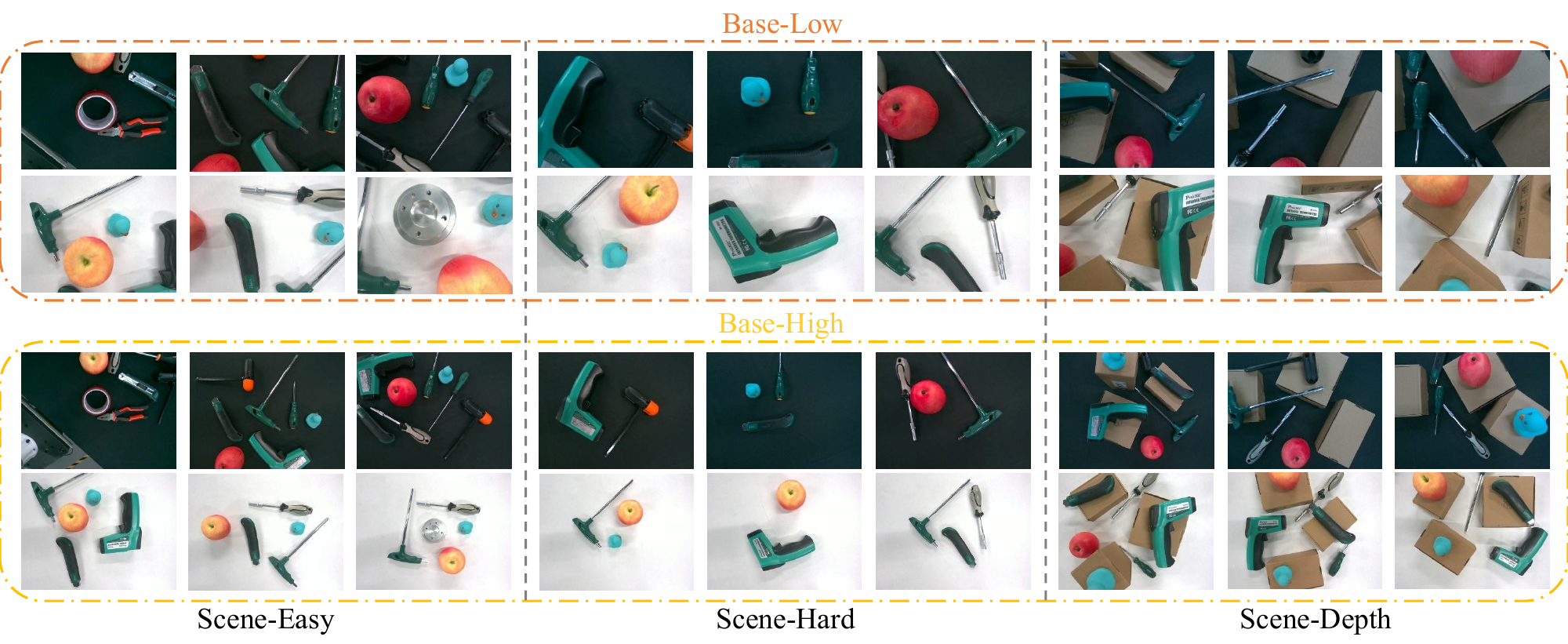}
    \captionsetup{skip=-3pt}
    \caption{Overview of varying degrees of proximity to the scene center for the robot's end-effector across three distinct environments. The upper images in each scene show the robot closer to the scene center (designated Base-Low), while the lower images show it farther away (referred to as Base-High).} 
    \label{fig:7_three_scenes}
\end{figure*} 

\begin{figure*}[htpb]        
    \centering 
    \vspace{-12pt}
    \includegraphics[scale=0.5]{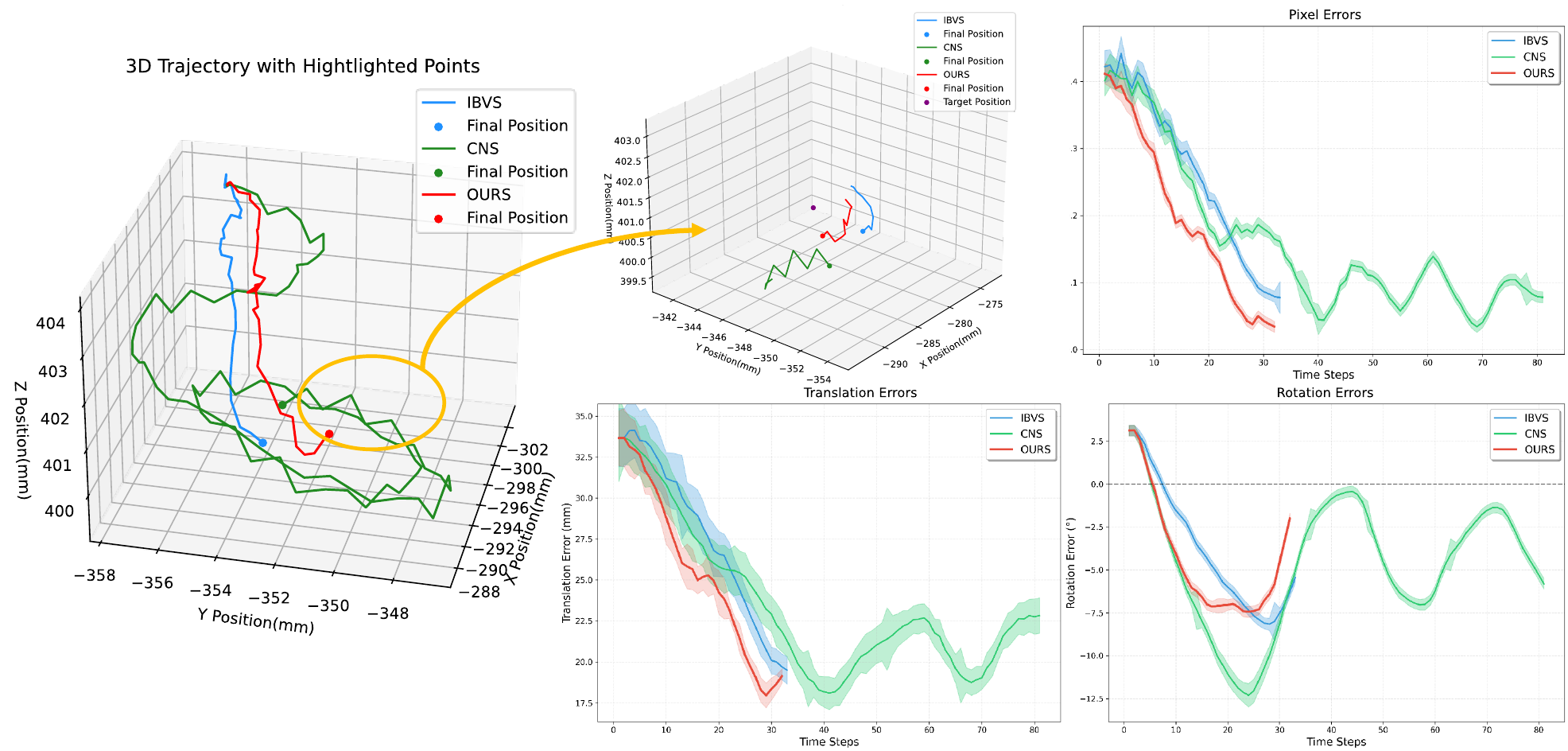}
    \captionsetup{skip=-2pt} 
    \caption{The images present the average comparison results of three methods across 10 runs under the challenging "Scene-Hard" setup. On the left, the illustration shows the overall trajectories of the three methods (all targeting the same goal) and a zoomed-in view of the final stage. Meanwhile, the three bottom-right images quantify pixel, translation, and rotation errors for the servoing task, tracking performance throughout the motion sequences. Colored shaded areas in the chart indicate their standard deviations.}
    \label{fig:8_errors}
\end{figure*} 

Our model's comprehension of spatial and geometric information has yielded improved outcomes in terms of translation error and convergence basin. Nevertheless, the image transformations from varied viewpoints can impact our understanding of the scene, particularly when depth information extends beyond the confines of the image boundaries.

\subsection{Real World Environment}

\begin{table*}[t]
    \centering
    \caption{Real-world experimental results comparing with IBVS, CNS, and Depth-PC quantify system performance in different feature extraction qualities, depth-variant scene adaptation, and complex depth perception capabilities. Within each level, we sample 50 pairs of initial and target poses, akin to the experimental setup in \cite{21}. }
    \label{tab:3_realworld}
    \resizebox{\textwidth}{!}{\begin{tabularx}{1.1\textwidth}{>{\hsize=1\hsize\centering\arraybackslash}X >{\centering\arraybackslash}X *{15}{c} }
        \toprule
        \multicolumn{1}{c}{Scene} & \multicolumn{3}{c}{Scene-Easy} & \multicolumn{3}{c}{Scene-Hard} & \multicolumn{3}{c}{Scene-Depth} & \multicolumn{3}{c}{Base-Low} & \multicolumn{3}{c}{Base-High}\\
        \cmidrule(lr){1-1} \cmidrule(lr){2-4} \cmidrule(lr){5-7}  \cmidrule(lr){8-10} \cmidrule(lr){11-13} \cmidrule(lr){14-16} 
        Method & IBVS & CNS & \textbf{Depth-PC} & IBVS & CNS & \textbf{Depth-PC} & IBVS & CNS & \textbf{Depth-PC} & IBVS & CNS & \textbf{Depth-PC} & IBVS & CNS & \textbf{Depth-PC}\\
        \midrule
        SR(\%) & 38 & 42 & \textbf{60} & 32 & 38 & \textbf{58} & 34 & 44 & \textbf{60} & 38.33 & 44 & \textbf{62.33} & 32 & 38.67 & \textbf{58.33}\\
        TE(mm) & 22.15 & 21.69 & \textbf{18.93} & 26.63 & 23.11 & \textbf{19.52} & 24.65 & 22.85 & \textbf{19.63} & 23.14 & 22.09 & \textbf{19.12} & 25.79 & 24.62 & \textbf{20.01}\\
        RE(\si{\degree}) & 3.65 & 3.28 & \textbf{2.01} & 7.02 & 5.96 & \textbf{2.34} & 5.39 & 4.78 & \textbf{2.34} & 4.11 & 3.59 & \textbf{2.12} & 7.35 & 6.12 & \textbf{4.33}\\
        mTT(s) & 6.05 & 6.69 & \textbf{4.11} & 8.01 & 8.39 & \textbf{4.95} & 7.42 & 6.59 & \textbf{4.98} & 6.77 & 6.64 & \textbf{5.21} & 9.52 & 8.98 & \textbf{7.01}\\
        \bottomrule
    \end{tabularx}}
    \vspace{-15pt}
\end{table*}

\begin{figure}[htbp]
    \centering
    \includegraphics[width=0.6\linewidth]{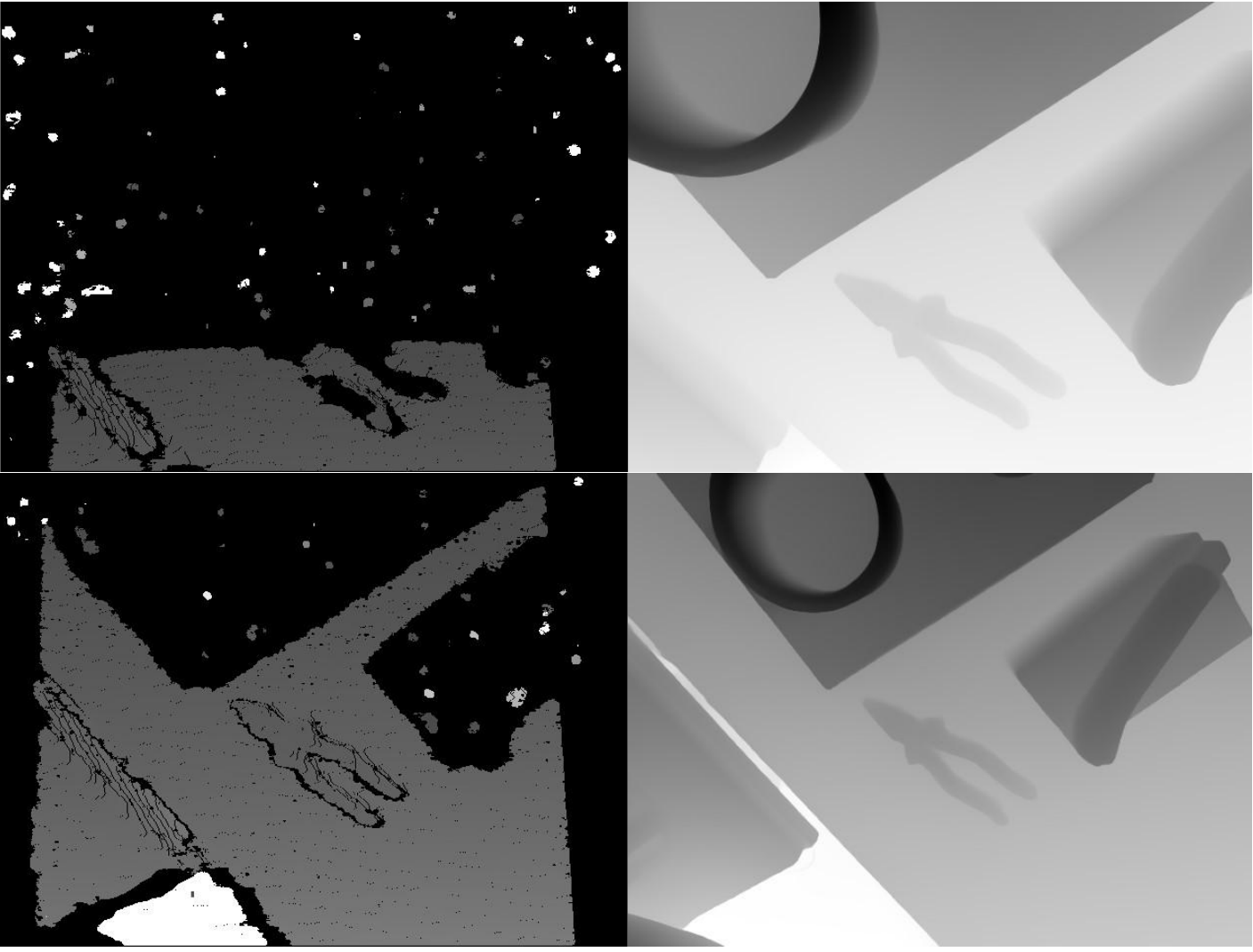}
    \caption{Visual mismatch between depth maps from the camera (left) and the depth estimator (right) for Base-Low (upper) and Base-High (lower) in Scene-Depth.}
    \label{fig:9_different_depths}
\end{figure}

We conducted comparative evaluations with CNS \cite{21} and IBVS \cite{1} in real-world. We designed three distinct scene settings to systematically investigate the robustness of our method under varying environmental complexities: (1). Scene-Easy incorporates \textbf{a sufficient number of objects to enhance keypoint visibility}, providing abundant matching references to simplify servo control; (2). Scene-Hard reduces the number of objects, leading to \textbf{scarce keypoints and increased matching difficulty}; (3). Scene-Depth introduces \textbf{objects at different depth levels}, a design addressing the inherent limitation of 3D cameras whose perceptual capabilities fluctuate across depth ranges as illustrated in Fig. \ref{fig:9_different_depths}. To further deepen the analysis, we also categorized tasks into Base-Low and Base-High based on the distance from the robot’s end-effector to the scene center, enabling comparative studies across scenes with varying distance configurations, as Fig. \ref{fig:7_three_scenes} illustrates. 

\textbf{Notion:} In our experiments, objects are randomly selected in the laboratory and have no relation with those in the training datasets. Considering the balance between performance and efficiency, we choose AKAZE as the feature detector and DepthAnythingv2-Small as the depth estimator. \textbf{Hardware}: Both training and inference are performed on a RTX 3090 GPU, with experimental validation conducted on a JK5 robot platform equipped with an Intel RealSense D435i camera.

As shown in TABLE \ref{tab:3_realworld}, our methodology achieves approximately 20\% higher SR with superior performance in complex scenes and depth-varying environments. Our model exhibits an enhanced ability to generalize across various settings, outperforming baselines in 3D geometric understanding and servo convergence—findings consistent with simulation results. Notably, while showing modest advantages in simulation, the method demonstrates a significant real-world performance leap, verifying its effectiveness in Sim2Real transfer. As shown in Fig. \ref{fig:8_errors}, the error curves demonstrate our method’s superiority in error minimization and convergence speed, along with its robustness to varying external noise and ability to generate smoother motion trajectories. Notably, IBVS achieves higher control frequency by only computing the Jacobian matrix, thus avoiding the increasing positional errors at the end, whereas our method and CNS—both relying on feature extractors and GNN—exhibit rising positional errors at the final stage.

\subsection{Ablation Study}

\noindent
\textbf{Simulation environment.} Initially, to assess the influence of data processing, we substitute the CNS method's original data handling with our refined approach. The results in TABLE \ref{tab:4_HPR} clearly demonstrate that our data processing techniques, when reducing the training cost, adeptly preserve the model's precision. The training duration, which initially spans more than 5 hours, sees a reduction of nearly 20\%.

\begin{table}[htbp]
    \centering
    \caption{Ablation experiments on data processing and HPR represents our approach. The statistics are derived from experiments carried out within \(E_\mathrm{YCB}\).}
    \begin{tabular}{ccccc}
        \toprule
            Method&SR(\%)&TE(mm)&RE(\(^{\circ}\))&TS(0.04s)\\  
        \midrule
            CNS\cite{21}+HPR & 99.33 & 1.603 & 0.132 & 200.7\\
            CNS & \textbf{100} & 1.592 & 0.134 & 207.8\\
            Depth-PC(w/o HPR)& 99.33 & 1.592 & 0.137 & 185.1\\
            Depth-PC(ours)&\textbf{100}&\textbf{1.426}&\textbf{0.124}&\textbf{178.5}
            \\
        \bottomrule
    \end{tabular}
    \label{tab:4_HPR}
\end{table}

To evaluate our Feature Fusion Module (FFM), we compare it with (Concat) and Cross Attention (CA) in TABLE \ref{tab:fusion}. Concatenation underperforms as it fails to capture relationships between positional and geometric features. CA improves performance slightly but incurs high computational cost due to pairwise attention calculations. In contrast, FFM enhances servo task performance while reducing resource usage: when adjusted to the same batch size, FFM’s GPU memory consumption is only 1/4 to 1/8 that of CA. This confirms FFM’s effectiveness for multimodal feature fusion.

\begin{table}[htbp]
    \centering
    \caption{Ablation experiments on feature fusion module. The statistics are derived from experiments carried out within \(E_\mathrm{YCB}\).}
    \label{tab:fusion}
    \begin{tabular}{ccccc}
        \toprule
            Method&SR(\%)&TE(mm)&RE(\(^{\circ}\))&TS(0.04s)\\  
        \midrule
            Depth-PC(Concat) & 92.67 & 2.656 & 0.366 & 293.4\\
            Depth-PC(CA)& 96.67 & 1.794 & 0.147 & 228.2\\
            Depth-PC(w/o DE)& 98.43 & 1.904 & 0.163 & 233.6\\
            Depth-PC(ours)&\textbf{100}&\textbf{1.426}&\textbf{0.124}&\textbf{178.5}
            \\
        \bottomrule
    \end{tabular}
\end{table}

Meanwhile, we have tested the performance of the model without Depth Embedding (w/o DE), which contributes a lot to convergence steps of the model. The model struggles in terms of all metrics in absence of it.

\noindent
\textbf{Real-world environment.} We then perform experiments based on sources of depth maps, evaluating these acquired from cameras and those processed by the depth estimator. 
\begin{table}[h]
    \centering
    \caption{The average results from ablation experiments on sources of depth maps.}
    \begin{tabular}{ccccc}
        \toprule
            Method&SR(\%)&TE(mm)&RE(\(^{\circ}\))&TS(s)\\  
        \midrule
            Depth-PC(Camera) & 42.32 & 27.067 & 6.456 & 9.856\\
            Depth-PC(ours)&\textbf{59.67}&\textbf{18.720}&\textbf{3.156}&\textbf{6.006}
            \\
        \bottomrule
    \end{tabular}
    \label{tab:6_different_depths}
\end{table}

It is clearly observable based on Fig. \ref{fig:9_different_depths} and TABLE \ref{tab:6_different_depths} that when the camera on the robot's end-effector is closer to the scene center, its interpretation of the environment's geometric details is less accurate and more heavily impacted by noise. In contrast, the depth estimator demonstrates a robust ability to understand the scene effectively.

\subsection{Limitations}
\begin{figure}[htbp]
    \vspace{-8pt}
    \centering
    \includegraphics[width=0.8\linewidth]{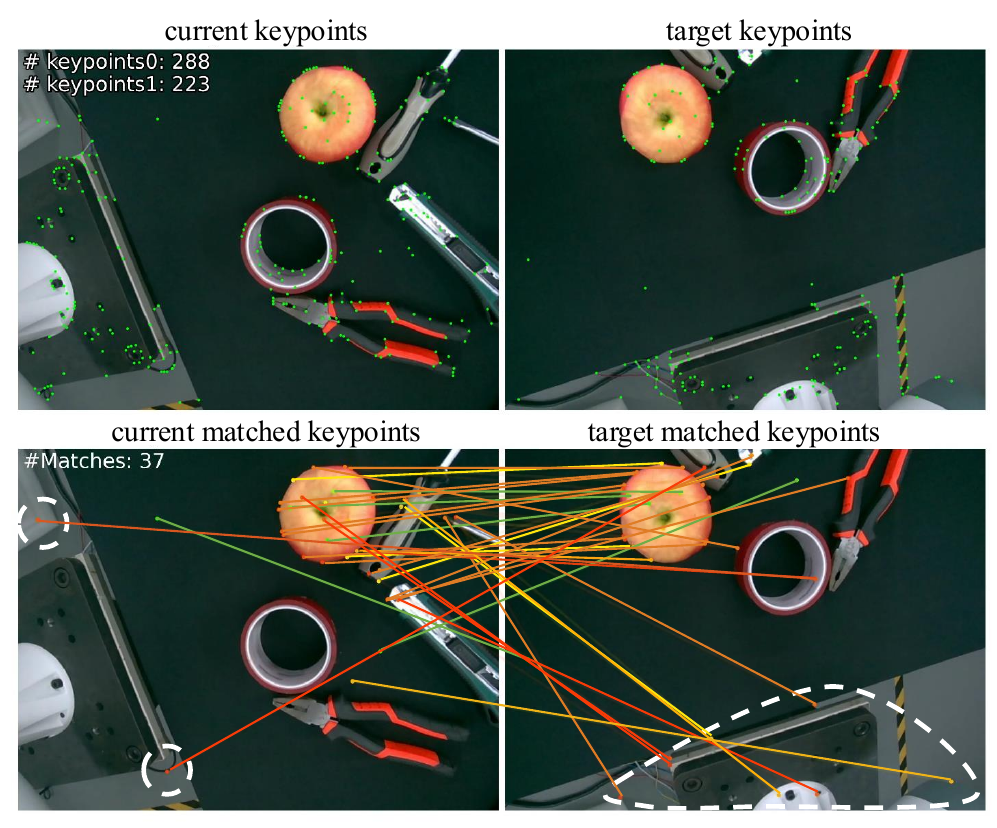}
    \caption{Visualization of noise points and matching status during the servo process.}
    \label{fig:noise}
\end{figure}

To enhance the model's generalization, we model the point cloud distribution in 3D space. However, there exists a domain gap between the one-to-one correspondence during training and keypoints correspondence obtained by 2D matching, which is exacerbated by issues such as mismatches caused by environmental noise in real-world settings—like failure cases shown in the white dashed area in Fig. \ref{fig:noise}. 

Additionally, our designed controller adopts distance-decoupled velocity to achieve zero-shot adaptation across diverse scenarios, yet the output velocity needs to be correlated with scene scale—a parameter that is difficult to accurately acquire during real-world robot motion. Meanwhile, relative depth input, which helps the model better perceive spatial knowledge, also exacerbates the aforementioned problem.

\section{CONCLUSIONS}

We develop a framework dedicated to servo tasks, which includes distinct processes for training and inference to achieve Sim2Real transfer in a zero-shot way. By integrating relative depth information, we facilitate cross-modal alignment and fusion between point clouds and the depth features of images. Leveraging feature detectors and depth estimators, we have narrowed the gap between the simulation environment and the real world, outperforming prior methods and validating its effectiveness in Sim2Real transfer. In addition, we have delved into the effective fusion of multimodal features in visual servoing, and have successfully attained commendable outcomes in both performance and training costs. Our model has demonstrated robust capabilities and remarkable generalization across diverse complex environments, as evidenced by real-world experimental results.

\vfill

\end{document}